\pdfoutput=1
\documentclass{article} 
\usepackage{not_clear,times}

\usepackage{amsmath,amsfonts,bm}

\def\eqref#1{equation~\ref{#1}}

\def\1{\bm{1}}

\DeclareMathAlphabet{\mathsfit}{\encodingdefault}{\sfdefault}{m}{sl}
\SetMathAlphabet{\mathsfit}{bold}{\encodingdefault}{\sfdefault}{bx}{n}

\usepackage{hyperref}       
\usepackage{url}            
\usepackage{booktabs}       
\usepackage{amsfonts}       
\usepackage{nicefrac}       
\usepackage{microtype}      
\usepackage{graphicx}
\usepackage{subcaption}
\usepackage{algorithm}
\usepackage{algpseudocode}

\usepackage{amsfonts}
\usepackage{color}

\newcommand{\eg}{{\it e.g.}}
\newcommand{\ie}{{\it i.e.}}

\newcommand{\reals}{{\mathbb{R}}}
\newcommand{\nats}{{\mathbb{N}}}

\newcommand{\Expect}[0]{\mathop{\mathbb{E}}}

\title{Strength in Numbers: Trading-off Robustness and Computation via Adversarially-Trained Ensembles}

\author{Edward Grefenstette \\
  DeepMind, London, UK \\
  \texttt{etg@google.com}\\
  \And
  Robert Stanforth\\
  DeepMind, London, UK\\
  \texttt{stanforth@google.com}\\
  \And
  Brendan O'Donoghue\\
  DeepMind, London, UK \\
  \texttt{bodonoghue@google.com}\\
  \And
  Jonathan Uesato\\
  DeepMind, London, UK \\
  \texttt{juesato@google.com}\\
  \And
  Grzegorz Swirszcz\\
  DeepMind, London, UK \\
  \texttt{swirszcz@google.com}\\
  \And
  Pushmeet Kohli\\
  DeepMind, London, UK \\
  \texttt{pushmeet@google.com}\\
}

\begin{document}

\maketitle

\begin{abstract}
While deep learning has led to remarkable results on a number of challenging problems, researchers have discovered a vulnerability of neural networks in adversarial settings, where small but carefully chosen perturbations to the input can make the models produce extremely inaccurate outputs. This makes these models particularly unsuitable for safety-critical application domains (\eg~self-driving cars) where robustness is extremely important. Recent work has shown that augmenting training with adversarially generated data provides some degree of robustness against test-time attacks. In this paper we investigate how this approach scales as we increase the computational budget given to the defender. We show that increasing the number of parameters in adversarially-trained models increases their robustness, and in particular that ensembling smaller models while adversarially training the entire ensemble as a single model is a more efficient way of spending said budget than simply using a larger single model. Crucially, we show that it is the adversarial training of the ensemble, rather than the ensembling of adversarially trained models, which provides robustness.
\end{abstract}

\section{Introduction}
Deep neural networks have demonstrated state-of-the-art performance in a wide range of application domains~\cite{krizhevsky2012imagenet}. However, researchers have discovered that deep networks are in some sense `brittle', in that small changes to their inputs can result in wildly different outputs~\citep{huang2017adversarial,jia2017adversarial,szegedy2013intriguing}. For instance, practically imperceptible (to human) modifications to images can result in misclassification of the image with high confidence. Not only are networks susceptible to these `attacks', but these attacks are also relatively easy to compute using standard optimization techniques~\citep{carlini2017towards,goodfellow2014explaining}. These changes are often referred to as \emph{adversarial perturbations}, in the sense that an adversary could craft a very small change to the input in order to create an undesirable outcome. This phenomenon is not unique to image classification, nor to particular network architectures, nor to particular training algorithms~\citep{papernot2016transferability, papernot2017practical}.

Adversarial attacks can be broken into different categories depending on how much knowledge of the underlying model the adversary has access to. In `white-box' attacks the adversary has full access to the model, and can perform both forward and backwards passes (though not change the weights or logic of the network)~\citep{carlini2017adversarial,goodfellow2014explaining}. In the `black-box' setting the adversary has no access to the model, but perhaps knows the dataset that the model was trained on ~\citep{papernot2016transferability,papernot2017practical}. Despite several recent papers demonstrating new defences against adversarial attacks~
\citep{akhtar2018threat, guo2017countering,liao2017defense,song2017pixeldefend,tramer2017ensemble,warde201611,xie2017mitigating,yuan2017adversarial}, recent papers have demonstrated that most of these new defences are still susceptible to attacks and largely just obfuscate the gradients that the attacker can follow, and that non-gradient based attacks are still effective~\cite{uesato2018adversarial, obfuscated-gradients}.

\paragraph{Exploring Tradeoff of Computation and Robustness}
In many safety-critical application domains (\eg~self-driving cars), robustness is extremely important even if it comes at the cost of increased computation. This motivated the central question considered by this paper:
\emph{Is it possible to increase adversarial robustness of a classifier at the cost of increased computation?}

There are a number of possibilities to employ extra computation available at runtime. We can use a much larger model that requires more time to run, execute the original model multiple times and aggregate the predictions, or instead of using a single model, make predictions from a portfolio or ensemble of models.
While researchers have proposed the use of portfolios and ensembles as a mechanism to improve adversarial robustness~\cite{AbbasiG17,Strauss17}, our experimental results indicate that stronger adversaries are able to attack the ensembles successfully.

\paragraph{Contributions}
In this paper, we study and analyze the trade-off of adversarial robustness and computation (memory and runtime). We propose the use of adversarial training of ensemble of models and through an  exhaustive ablative analysis make the following empirical findings:
\begin{itemize}
\item increased computation and/or model size can be used to increase robustness,
\item ensembles on their own are not very robust, but can be made robust through  adversarial training where the ensemble is treated as a single model,
\item adversarially trained ensembles are more robust than adversarially trained individual models requiring the same amount of parameters/computation
\end{itemize}

\paragraph{Related Work}
Recently,~\citet{tramer2017ensemble} investigated the use of ensembles for adversarial robustness. However, their goal and approach was quite different from the technique we are investigating. In~\citet{tramer2017ensemble}, the authors generated adversarial perturbations using an ensemble of \emph{pre-trained} models in order to transfer the example to another model during training. This procedure decouples adversarial example generation from the current model, and consequently the model being trained cannot simply `overfit' to the procedure for generating adversarial examples, which they generally took to be single-step attack methods. The authors demonstrated strong robustness of the resulting trained model to black-box attacks. By contrast, in this paper we investigate using an ensemble of models as our predictive model, and we train the models using multi-step adversarial training. We show increased robustness to both black-box and white-box adversarial attacks using this strategy.

\section{Preliminaries}
Here we lay out the basics of attacking a neural network by the generation of adversarial examples.
Denote an input to the network as $x \in \reals^d$ with correct label $\hat y \in \mathcal{Y} \subset \nats$, and let $m_\theta : \reals^d \rightarrow \reals^{|\mathcal{Y}|}$ be the mapping performed by the neural network which is parameterized by $\theta \in \reals^p$. Let $L: \mathcal{Y}
\times \reals^{|\mathcal{Y}|} \rightarrow \reals$ denote the loss we are trying to minimize (\eg, the cross-entropy). When training a neural network we seek to solve
\begin{equation}
\label{e-loss}
\begin{array}{ll}
\mbox{minimize} & \Expect_{(x, y) \sim \mathcal{D}} L(\hat y, m_\theta(x))
\end{array}
\end{equation}
over variable $\theta$, where $\mathcal{D}$ is the data distribution. Given any fixed $\theta$ we can generate (untargeted) adversarial inputs by perturbing the input $x$ so as to \emph{maximize} the loss. We restrict ourselves to small perturbations around a nominal input, and we denote by $\mathcal{B}$ this set of allowable inputs. For example, if we restrict ourselves to small perturbations in $\ell_\infty$ norm around a nominal input $x^\mathrm{nom}$ then we could set $\mathcal{B} = \{x \mid \|x - x^\mathrm{nom}\|_\infty \leq \epsilon \}$ where $\epsilon > 0$ is the tolerance. A common approach for generating adversarial examples is projected gradient descent~\cite{carlini2016defensive}, \ie, to iteratively update the input $x$ by
\begin{equation}
\label{e-pgd}
\tilde x^{k+1} = \Pi_\mathcal{B} (\tilde x^k + \eta \nabla_x L(y, m_\theta(\tilde x^k))),
\end{equation}
where typically $x^0 = x + \epsilon$ for some noise $\epsilon$, $\eta > 0$ is a step-size parameter and $\Pi_\mathcal{B}$ denotes the Euclidean projection on $\mathcal{B}$. We add noise to the initial point so that the network can't memorize the training dataset and mask or obfuscate the gradients at that point~\cite{uesato2018adversarial, obfuscated-gradients}, in other words the added noise encourages generalization of adversarial robustness to the test dataset. If instead of using the gradient we just use the sign of the gradient then this is the fast-gradient-sign method~\cite{goodfellow2014explaining}.
Empirically speaking, for most networks just a few steps of either of these procedures is sufficient to generate an $\tilde x$ that is close to $x^\mathrm{nom}$ but has a different label with high confidence.

In this paper we are primarily concerned with the performance of ensembles of models when trained with \emph{adversarial training}~\cite{madry2017towards}. In adversarial training we train a network to minimize a weighted sum of two losses (where the relative weighting is a hyper-parameter). The first loss is the standard loss of the problem we are trying to solve on the normal training data, \eg, the cross-entropy for a classification task. The second loss is the same function as the first loss, except evaluated on \emph{adversarially generated data}, where typically the adversarial data is generated by attacking the network at that time-step. In other words we replace the problem in eq.~(\ref{e-loss}) with
\begin{equation}
\label{e-adv-lossfn}
\begin{array}{ll}
\mbox{minimize} & \Expect_{(x, y) \sim \mathcal{D}} (L(\hat y, m_\theta(x)) + \rho L(\hat y, m_\theta(\tilde x)))
\end{array}
\end{equation}
where $\rho \geq 0$ is the weighting parameter and $\tilde x$ is an adversarial example generated from $x$ at model parameters $\theta$ using, for example, the update in eq.~(\ref{e-pgd}). This problem is usually approximated by sampling and minimizing the empirical expectation.

\section{Adversarially-trained Ensembles}

In this section we lay out the basic strategy of using ensembles of models to increase robustness to adversarial attacks. The notion of ensemble used here simply involves taking $k$ separately-parameterized models and averaging their predictions. If the output of network $i$ as a function of input $x$ and with network parameters $\theta_i$ is given by $p(\cdot | x, \theta_i) = m_{\theta_i}(x)$, then the output of the ensemble is
\[
    p(y|x) = \frac{1}{k}\sum_{i=1}^k{p(y|x, \theta_i)}.
\]
Alternatively, we could consider using a `gating network' to generate data-dependent weights for each model rather than a simple average, though we found the performance to be similar.

Using ensembles to improve the performance of statistical models is a very old idea; see, \eg~\citet{opitz1999popular} for a survey. The basic intuition is that several weak models can be combined in such a way that the ensemble performs better than any individual, and is sometimes explained as being caused by the errors of the models `cancelling' with one another.

In order to ensure that the models are actually producing different outputs the diversity of the models must be maintained. This can be done in several ways, such as bootstrapping the data, whereby each model gets a slightly different copy of the data, or using totally different model types or architectures. In the case that the model training procedure is convex, and if all models architectures are the same and are getting the same data, then the models in the ensemble would be expected to converge on the same parameters. In the case of neural networks however, the model training procedure is not convex and so our strategy for maintaining diversity is very simple---initialize each model differently. Due to the nature of training neural networks it is likely that differently initialized networks will converge (assuming they do, in fact, converge) to different points of the parameter space. The insight that only different initialization is required is not new, previous papers have observed that different initialization is sufficient for uncertainty estimation~\cite{lakshminarayanan2016simple, osband2016deep}.

Different initialization for networks has an appealing interpretation. If we take a Bayesian approach to the classification problem, then we have a prior over possible model parameters, $p(\theta)$, a likelihood of the data, $p(D | \theta)$, and a probability of a label $y$ given an input and a model, $p(y |x, \theta)$. The `Bayes-optimal' classification of a new data point $x$ is given by
\[
y^\star  = \mathrm{argmax}_{y} \int_\theta p(y | x, \theta) p(D | \theta) p(\theta).
\]
This classifier is optimal in the sense that no other classifier can outperform it on average, given the same model class and knowledge of the prior and likelihood; however, the formulation is intractable for all but small problems. We can consider approximating it by the following approach, sample initial parameters from the prior $p(\theta)$ and run an iterative procedure to (approximately) maximize the likelihood $p(D|\theta)$. Very loosely speaking, we can consider this procedure as approximately sampling from the posterior over models $p(\theta | D) \propto p(D | \theta) p(\theta)$. Consequently, we output the classification
\[
y^\star  = \mathrm{argmax}_{y} \sum_{i=1}^k p(y | x, \theta_i),
\]
\ie, the best guess of the ensemble. The role of initialization therefore is that of sampling from our prior over possible model parameters.

\paragraph{Adversarial training of Ensembles} Up to this point we have discussed the use of ensembles for improving classification performance and approximating the Bayes optimal classifier. Typically speaking neural networks appear to not benefit much from ensembling in terms of nominal performance. Here, however, we make the claim that adversarially trained ensembles of networks provide a level of robustness to adversarial attacks.
When using ensembles the loss function for adversarial training in (\ref{e-adv-lossfn}) is replaced by the mean of the loss over the $k$ models, \ie, now we want to solve
\begin{equation}
\label{e-adv-lossfn-ens}
\begin{array}{ll}
\mbox{minimize} & \Expect_{(x, y) \sim \mathcal{D}} \left(L(\hat y, \frac{1}{k} \sum_{i=1}^k m_{\theta_i}(x)) + \rho L(\hat y, \frac{1}{k} \sum_{i=1}^k m_{\theta_i}(\tilde x))\right)
\end{array}
\end{equation}
over variables $\theta_i$, $i=1,\ldots,k$, and where $\tilde x$ is an adversarial example generating by attacking the entire ensemble. The exact procedure is outlined in Algorithm \ref{a-adv-train}.  We demonstrate empirically in the numerical results section that this procedure increases robustness to adversarial inputs. Following these results, we offer an analysis and hypothesis why ensembles outperform single models, even when controlling for number of parameters.

\begin{algorithm}[t]
\caption{Adversarial ensemble training using PGD under $\ell_\infty$ norm constraint}
\begin{algorithmic}
\State {\bf input:} $k$ neural networks $m_{\theta_i}$, $i=1, \ldots, k$;
attack steps $N$; step sizes $\eta$, $\hat \eta$;
initial variance $\sigma$; adversarial loss weighting $\rho$; perturbation width $\delta$
\State {\bf initialize:} neural network parameters $\theta_i^0$ randomly, $i=1, \ldots, k$
\For{time-step $t=0,1,\ldots,$}
\State sample input minibatch $(x, \hat y) \sim \mathcal{D}$
\State initialize $\tilde x^0 = x + \epsilon$ where $\epsilon \sim \mathcal{N}(0, \sigma^2 I)$
\State define $\mathcal{B} = \{x^\prime \mid \|x - x^\prime\|_\infty \leq \delta\}$
\For{$k = 0, \ldots, N - 1$}
\[
\tilde x^{k+1} = \Pi_\mathcal{B} (\tilde x^k + \hat \eta \nabla_x L(\hat y, \frac{1}{k} \sum_{i=1}^k m_{\theta_i}(\tilde x^k)))
\]
\EndFor
\State update parameters for each $i=1,\ldots, k$:
\[
\theta_i^{t+1} = \theta_i^t - \eta \nabla_{\theta_i} \left( L(\hat y, \frac{1}{k} \sum_{j=1}^k m_{\theta_j^t}(x)) + \rho L(\hat y, \frac{1}{k} \sum_{j=1}^k m_{\theta_j^t}(\tilde x^N)) \right )
\]
\EndFor
\end{algorithmic}
\label{a-adv-train}
\end{algorithm}

\section{Experimental Setup}

\begin{figure}[t!]
    \centering
    \includegraphics[width=.9\linewidth]{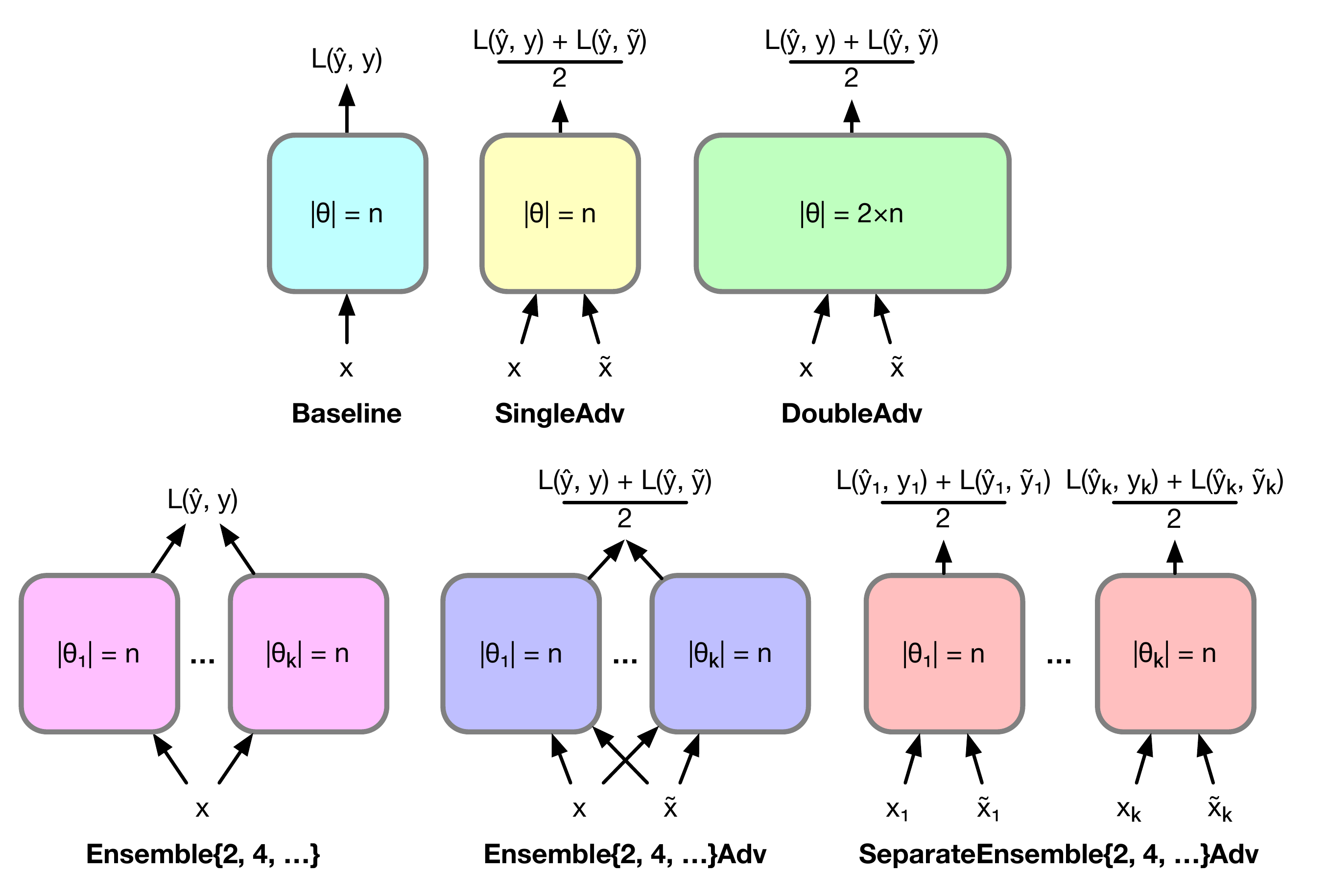}
    \caption{Schematic depiction of classes of models compared in this paper. Here, $n$ indicates the number of parameters in the base model,  $\hat{y}$ indicates the ground trouth label, $x$ is a clean input from the dataset, $\tilde{x}$ is that input after a number of steps of the chosen adversarial training attack (7 steps of IFGSM in our experiments), $y$ is the output distribution according to the network based on clean input $x$, and $\tilde{y}$ is the output based on adversarial input $\tilde{x}$. Adversarially trained networks are shown to have two inputs (and two losses) for compactness, but in practice two parameter-sharing copies of the network will be instantiated, with one taking clean input, the other taking adversarial input, and their losses will be computed separately and averaged before optimisation.}
    \label{fig:models}
\end{figure}

\subsection{Models Compared}

\paragraph{Non-Adversarial Benchmarks} The \textbf{Baseline} model for our investigation is a Wide ResNet~\citep{zagoruyko2016wide} consisting of a $3 \times 3$ convolution layer, followed by three layers containing 28 ResNet blocks of width factor 10, followed by batch normalization~\cite{ioffe2015batch} layer, followed by a ReLU~\citep{nair2010rectified}, and by a final linear layer projecting into the logits of the CIFAR-10 classes. All models we experimented with here are variations on this architecture, and where hyperparameters are not explicitly referenced, they are assumed to be the same as this base model. \textbf{Ensemble2} contains two copies of the baseline architecture. This has twice the number of parameters of the baseline. Together with the base model, these constitute our non-adversarially trained benchmarks.

\paragraph{Adversarial Models} When adding adversarial training to the baseline architecture, we obtain our \textbf{SingleAdv} benchmark, which has the same number of parameters as the baseline. When trained with adversarial training, whereby the whole ensemble is attacked by Iterated Fast Gradient Sign Method (IFGSM)~\citep{kurakin2016adversarial} at each training step to obtain adversarial inputs, we refer to the ensemble as \textbf{Ensemble2Adv}. This ensemble has as many parameters as its non-adversarially-trained counterparts.

\paragraph{Comparisons to Ensemble2Adv} In order to compare \textbf{Ensemble2Adv} to the \textbf{SingleAdv} benchmark while controlling for number of parameters, we introduce a variant \textbf{DoubleAdv} of this benchmark with ResNet blocks of width 15, which yields roughly the same number of parameters as \textbf{Ensemble2Adv}. Finally, we train two separately parameterised instances of \textbf{SingleAdv} and ensemble them at test time for the purpose of evaluating the hypothesis that it is adversarial training of ensembles that provides and advantage, and call this test-time model \textbf{SeparateEnsemble2Adv}.

The model variations described here are illustrated in Figure~\ref{fig:models}, which can serve as a basis for repeating these experiments with a different base model architecture.

\subsection{Training Procedure}

We train and evaluate our models on CIFAR-10~\citep{krizhevsky2009learning}. We use similar hyperparameters to~\cite{zagoruyko2016wide}, with additional iterations to account for the fact that minimizing the adversarial objective requires more training steps.  We train all models for 500,000 iterations using a momentum optimiser with minibatches of size 128, with an initial learning rate of 0.1, a momentum of 0.9, and a learning rate factor (decay) of 0.2 after $\{30\textrm{k}, 60\textrm{k}, 90\textrm{k}\}$ steps. When doing adversarial training, we train both on ``clean'' versions of the minibatch images, and on adversarial examples produced by $7$ steps of IFGSM, following~\cite{madry2017towards}. The cross-entropy losses with regard to the ground truth labels for both the adversarial and clean images are averaged to obtain gradients for the model (\ie~$\rho = 1$).

\subsection{Evaluation Procedure}

During training, we run an evaluation job which evaluates the accuracy of the model on the entire CIFAR-10 test set. We consider two white-box adversaries, both with a maximum $L_\infty$ perturbation of $8$ (out of 255): \textbf{IFGSM} which performs the iterated fast gradient sign method update, which is equivalent to steepest descent with respect to the $L_{\inf}$ norm~\cite{madry2017towards, kurakin2016adversarial} and \textbf{PGD} which performs projected gradient descent using the Adam~\cite{kingma2014adam} update rule. During training, we evaluate using \textbf{IFGSM7}, the training adversary which performs 7 iterations of the IFGSM update, also used in~\cite{madry2017towards}, as well as \textbf{PGD5} and \textbf{PGD20}, the 5 and 20-step versions of our PGD attack. Additionally, for the best model, we run these attacks 500 steps in order to estimate the strongest possible attacks.

We further include a black-box adversary in our evaluation procedure. We use a dataset of precomputed adversarial examples, following the procedure in~\cite{liu2016improved} against an ensemble of a Wide ResNet~\cite{zagoruyko2016wide} and VGG-like~\cite{simonyan2014very} architectures. The two models are trained with standard training procedures and achieve 96.0\% and 94.5\% accuracy respectively on the CIFAR-10 clean test set, and are ensembled by an arithmetic mean of their logits. The adversary is the \textbf{PGD20} adversary which fools all members of the ensemble on 100\% of the evaluation set. We note that the exact values for robustness of networks to black box attacks can be highly contingent on the similarity between the original and attacked networks~\cite{uesato2018adversarial}, rather than the true adversarial robustness of the attacked network. However, we include black box accuracies for best practice, as a check against models which achieve illusory robustness through obscured gradients~\cite{goodfellow2014explaining}.

We trained and evaluated each model with three separate random seeds. Evaluation outliers, caused by occasional crashes of evaluation jobs, are removed according to the following procedure. We compute a smoothed version of each time series by using a centered rolling median window of width 50. We take the absolute difference of each original time series and its smoothed form, compute the mean of the difference, and replace points in the original time series with their smoothed version only when the absolute difference exceeds three standard deviations with this mean. This removes at most two outlier points per model per evaluation in our runs. Evaluation time series for different seeds are then interpolated to obtain results on the same 1000 time-steps, which are then averaged across seeds, per model class.

\section{Results and Analysis}

\begin{table}[t]
    \centering

    \caption{Evaluation Results}
    \label{tab:eval_tables}

    \begin{subtable}[]{\linewidth}
    \centering
    \caption{Average of last 10 evaluation steps}
    \label{tab:last_10_eval}
    \begin{tabular}{lrrrrrr}
    \toprule
     & \textbf{\shortstack[l]{clean\\ accuracy}} &  \textbf{\shortstack[l]{IFGSM5\\ accuracy}}  &  \textbf{\shortstack[l]{PGD5\\ accuracy}} &  \textbf{\shortstack[l]{PGD20\\ accuracy}} &  \textbf{\shortstack[l]{black box\\ accuracy}} \\
    \midrule
         \textbf{Baseline} &            \textbf{0.94} &            0.34 &           0.15 &            0.01 &                0.27 \\
    \textbf{Ensemble2} &            \textbf{0.94} &            0.59 &           0.44 &            0.30 &                0.22 \\
    \textbf{Ensemble4} &            0.91 &            0.50 &           0.40 &            0.34 &                0.26 \\
    \textbf{SingleAdv} &            0.82 &            0.55 &           0.44 &            0.43 &                0.80 \\
    \textbf{DoubleAdv} &            0.83 &            0.57 &           0.46 &            0.44 &                0.82 \\
 \textbf{Ensemble2Adv} &            0.85 &            0.62 &           0.55 &            0.52 &                0.83 \\
 \textbf{Ensemble4Adv} &            0.87 &            \textbf{0.66} &           \textbf{0.58} &            \textbf{0.53} &                \textbf{0.85} \\
    \bottomrule
    \end{tabular}
    \end{subtable}

    \begin{subtable}[]{\linewidth}
    \centering
    \caption{Evaluation results for model at best IFGSM5 training step}
    \label{tab:best_eval}
    \begin{tabular}{lrrrrrr}
    \toprule
    & \textbf{\shortstack[l]{clean\\ accuracy}} &  \textbf{\shortstack[l]{FGSM5\\ accuracy}}  &  \textbf{\shortstack[l]{PGD5\\ accuracy}} &  \textbf{\shortstack[l]{PGD20\\ accuracy}} &  \textbf{\shortstack[l]{black box\\ accuracy}} \\
    \midrule
    \textbf{Baseline} &            \textbf{0.95} &            0.57 &           0.29 &            0.09 &                0.26 \\
    \textbf{Ensemble2} &            \textbf{0.95} &            0.65 &           0.52 &            0.38 &                0.22 \\
    \textbf{Ensemble4} &            0.93 &            0.60 &           0.48 &            0.43 &                0.24 \\
    \textbf{SingleAdv} &            0.84 &            0.57 &           0.46 &            0.45 &                0.81 \\
    \textbf{DoubleAdv} &            0.85 &            0.60 &           0.48 &            0.47 &                0.84 \\
 \textbf{Ensemble2Adv} &            0.87 &            0.64 &           0.56 &            \textbf{0.52} &                0.85 \\
 \textbf{Ensemble4Adv} &            0.88 &            \textbf{0.67} &           \textbf{0.58} &            \textbf{0.52} &                \textbf{0.86} \\
\bottomrule
    \end{tabular}
    \end{subtable}

  \begin{subtable}{\linewidth}
    \centering
    \caption{Model accuracy after 500 attack steps.}
    \label{tab:pgd_500}
    \begin{tabular}{lllllll}
    \toprule
    & & & & \multicolumn{2}{c}{\textbf{Ensemble}} & \\
    & \textbf{Baseline} & \textbf{SingleAdv} & \textbf{DoubleAdv} & \textbf{-2} & \textbf{-2Adv} & \textbf{Separate2Adv} \\
    \midrule
\textbf{IFGSM} & 0.16 & 0.46 & 0.47 & 0.13 & \textbf{0.55} & 0.49 \\
\textbf{PGD}  & 0.04 & 0.44 & 0.47 & 0.02  & \textbf{0.52} & 0.47 \\ \bottomrule
\end{tabular}
\end{subtable}
\end{table}

\begin{figure}[t]
\centering

\includegraphics[width=.9\linewidth]{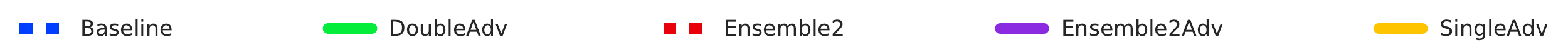}

\begin{subfigure}[b]{.48\linewidth}
\includegraphics[width=\linewidth,trim={2.7cm 0 2.9cm 0},clip]{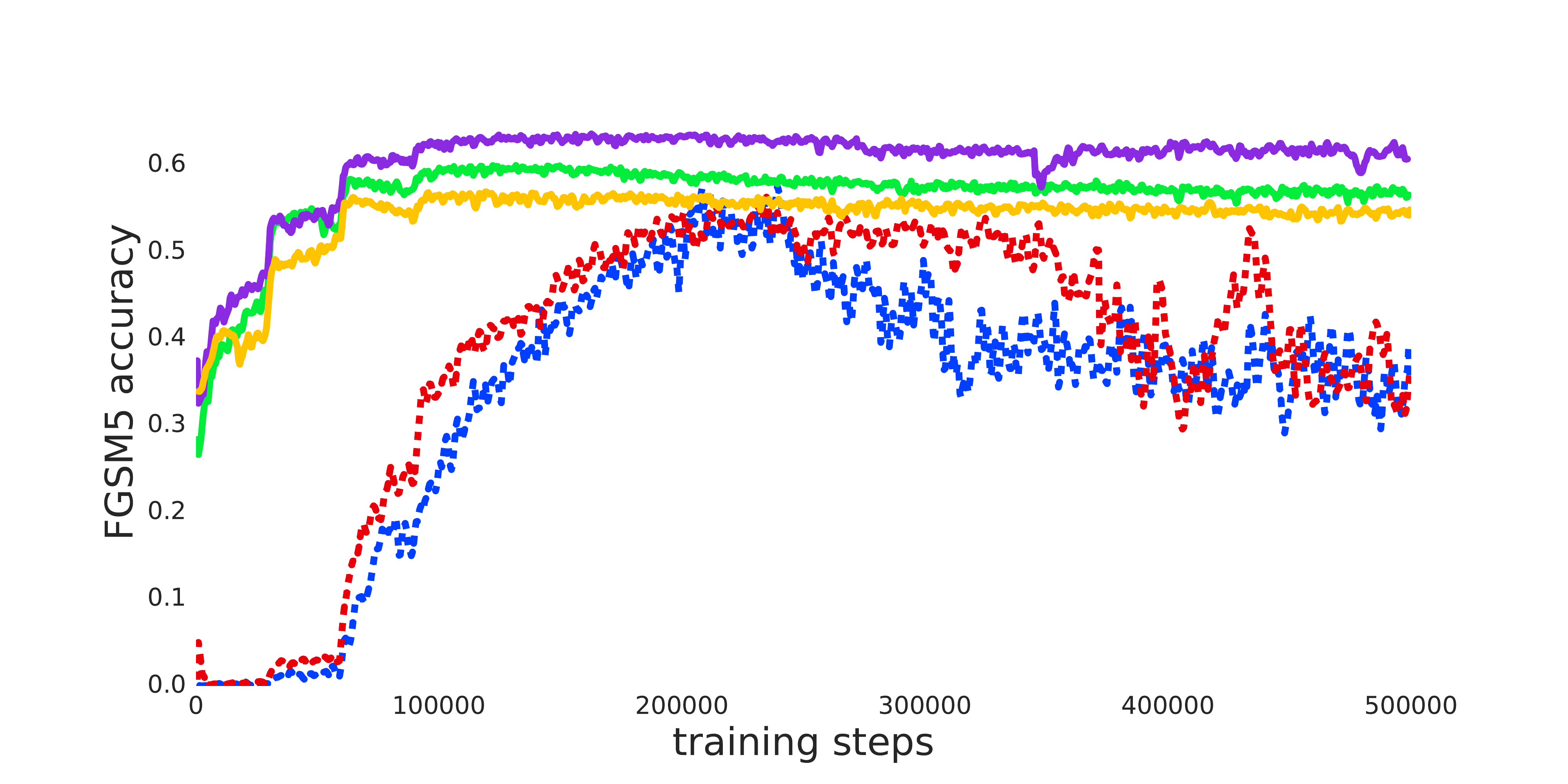}
\caption{Evaluation scores over training steps, 5 steps of IFGSM per evaluation.}\label{fig:fgsm}
\end{subfigure}
\hspace{.02\linewidth}
\begin{subfigure}[b]{.48\linewidth}
\includegraphics[width=\linewidth,trim={2.7cm 0 2.9cm 0},clip]{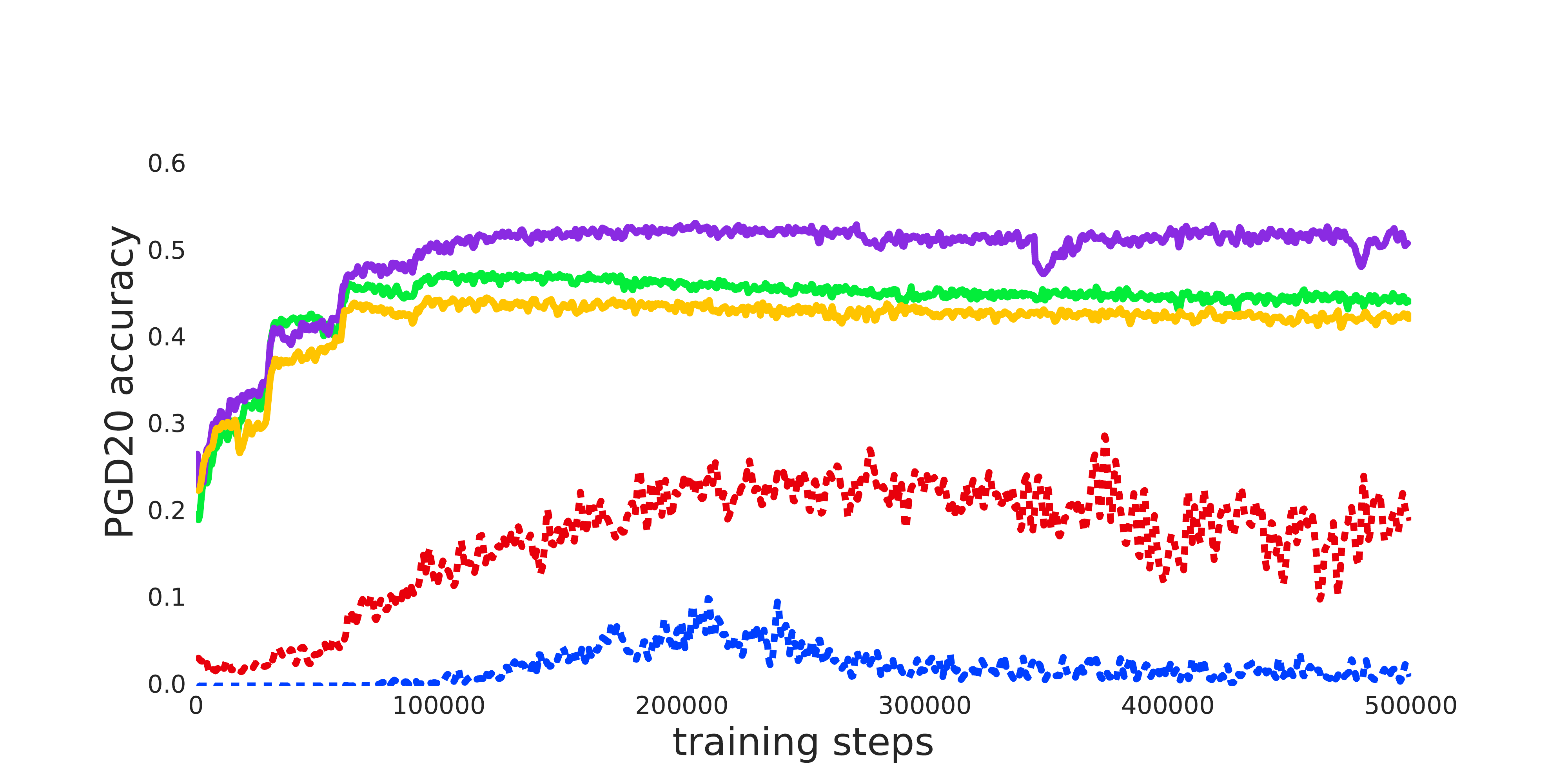}
\caption{Evaluation scores over training steps, 20 steps of PGD per evaluation.}\label{fig:pgd20}
\end{subfigure}
\begin{subfigure}[b]{\linewidth}
    \centering
    \includegraphics[width=0.9\linewidth,trim={.5cm .5cm 0cm .5cm},clip]{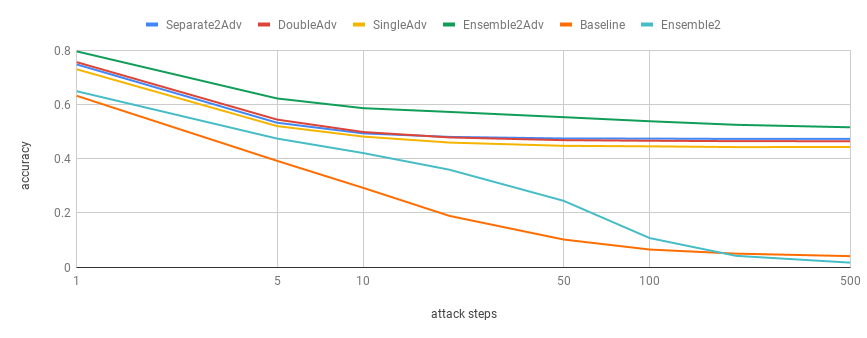}
    \caption{Accuracy under PGD attack as a function of the number of attack steps.}
    \label{fig:cw_over_time}
\end{subfigure}
\caption{Evaluation Curves}
\label{fig:eval_curves}
\end{figure}

We give a numerical break down of evaluation accuracies for the metrics described above, during and at the end training, in Table~\ref{tab:eval_tables}: in Table~\ref{tab:last_10_eval}, we report the average of the last 10 evaluation steps for all models, and in Table~\ref{tab:best_eval}, we report the evaluation metrics at the time step where each model obtained the best evaluation score on \textbf{FGSM5}. In Figures~\ref{fig:fgsm} and~\ref{fig:pgd20}, we show the evolution of evaluation accuracies for selected metrics. To more thoroughly evaluate the models compared here, we show in Figure~\ref{fig:cw_over_time} how the accuracy of our models drops as the number of PGD attack steps increases. We report the evaluation results for 500 steps of PGD of the model snapshots used for Table~\ref{tab:best_eval} in Table~\ref{tab:pgd_500}.

Figures~\ref{fig:fgsm} and~\ref{fig:pgd20} show that adversarially trained models uniformly outperform non-adversarially trained ones. Especially with weaker attacks, such as \textbf{IFGSM5} and \textbf{PGD5}, non-adversarially trained models exhibit some recovery of robustness to attacks after 2--300,000 steps of training, but this is not stable and decays with further training. We further confirm that even such models which achieve some robustness against weak adversaries have true adversarial robustness close to 0\% when the adversarial optimization is run for longer. In contrast, the robustness of adversarially trained models is stable throughout training. We read, in Table~\ref{tab:best_eval}, that all models incorporating adversarial training do slightly worse on the CIFAR-10 test, suffering a drop of roughly 10 points in accuracy, a phenomenon which was also observed in other work~\cite{madry2017towards}. On \textbf{PGD20}, the smallest gap between an adversarially trained model and a baseline is 22\%. \textbf{Ensemble2Adv} yields an improvement of 7\% over a \textbf{SingleAdv}, of 5\% over the parameterically equivalent \textbf{DoubleAdv}, and of 29\% over the non-adversarially trained \textbf{Ensemble2Adv}.

In Figure~\ref{fig:cw_over_time}, we see that while the accuracies of the \textbf{Ensemble2Adv} drop more readily as the number of attack steps increases, they preserve a gap 7 accuracy points over the \textbf{SingleAdv} benchmark. Here, we also compare to an ensemble, \textbf{Separate2Adv}, where the individual models in the ensemble were separately adversarially trained. We observe that this ensemble produces a robustness to adversarial attacks which is closer to the \textbf{SingleAdv} results than to \textbf{Ensemble2Adv}, despite having the exact same structure and number of parameters. We present the evaluation accuracies after 500 steps of PGD in Table~\ref{tab:pgd_500}, which maintains the relative ordering and rough gaps between models seen in Table~\ref{tab:best_eval}, thereby helping validate our results.

\section{Conclusions and Further Work}

In this paper, we provide an empirical study of the effect of increasing the number of parameters in a model trained with adversarial training methods, with regard to its robustness to test-time adversarial attacks. We showed that while increasing parameters improves robustness, it is better to do so by ensembling smaller models than by producing one larger model. Through our experiments, we show that this result is not only due to ensembling alone, or to the implicit robustness of an ensemble of adversarially trained models, but specifically to due to the adversarial training of an ensemble as if it were a single model. Further work should seek to determine whether scaling the number of models in the ensemble while controlling for number of parameters produces significant improvements over the minimal ensembles studied here in an attempt to draw conclusions about why such architectures are generally more robust than larger single models, even under adversarial training.


\bibliographystyle{iclr2019_conference}
\bibliography{references}

\appendix
\newpage
\section{Additional Evaluation}

We performed a number of additional evaluation experiments to increase our confidence in evaluating against the strongest possible attacks.
In these evaluations, a common concern is that because the proposed ensembling mechanism involves applying a softmax within each individual model before averaging the probabilities, this softmax could mask gradients due to saturating the softmax \citep{carlini2017towards}.

To address this concern, we run a number of experiments which remove this softmax.
More concretely, denote the logits of model $k$ as $z = Z(x, \theta_k)$, so that $p(y|x, \theta_k) = softmax(Z(x, \theta_k))$ where both $p(y)$ and $z$ are represented by vectors with shape matching the number of possible labels.
Then, following \citet{carlini2017towards}, the loss function we use in each adversary is
$$Z(x, \theta_k)_t - \max_{i \neq t} Z(x, \theta_k)_i$$
where $t$ is the correct label. This loss is less than $0$ if and only if $x$ is misclassified by model $k$.

We then perform projected gradient descent on this objective. We continued doubling the number of iterations until the accuracy under attack stopped decreasing.

\subsection{Submodel Evaluation}

Using our best model, \textbf{Ensemble4Adv}, we evaluated each model in the ensemble separately. We attack each model using the procedure described above.
We note that this evaluation procedure very closes matches \citet{madry2017towards}, in that gradients are taken with respect to the same model architecture (a Wide ResNet without the final softmax) trained using similar training procedures (adversarial training).
Empirically, PGD performs nearly as well as the best adversaries against the model used by \citet{madry2017towards}, so we might expect it to perform similarly against the model evaluated here.
Figure \ref{fig:appendix_plots} shows results against PGD with increasing numbers of iterations - we note that accuracy plateaus after 100 iterations.

Table \ref{tab:eval_tables_extra} reports accuracy against our strongest adversary.
Interestingly, the most robust single model has $43.9\%$ adversarial accuracy, slightly lower than the model in \citet{madry2017towards}, and significantly less than the best attack against the entire ensemble. Further, many of the other models have adversarial accuracy significantly lower than this.

One possible explanation is that individual models tend to have higher confidences on the images on which they are adversarially robust, and thus receive a larger weight in the ensemble on their adversarially robust predictions, compared to their incorrect predictions. We have not investigated this hypothesis, but believe it is an interesting direction for future research. We cannot completely rule out the possibility that a stronger adversary could further reduce the accuracy of the full ensemble, but we have made a best effort to evaluate the ensemble against the strongest possible attacks, which we describe in the section below.

\begin{table}[h]
    \centering
    \caption{Evaluation against strongest adversary}
    \label{tab:eval_tables_extra}
    \begin{tabular}{lrr}
    \toprule
     & \textbf{\shortstack[l]{clean\\ accuracy}} &  \textbf{\shortstack[l]{adversarial\\ accuracy}} \\
    \midrule
    \textbf{Model 1} &           63.8 &            34.4  \\
    \textbf{Model 2} &           79.1 &            43.9  \\
    \textbf{Model 3} &           69.1 &            38.7  \\
    \textbf{Model 4} &           49.0 &            23.7  \\
    \midrule
    \textbf{Ensemble} &         87.6 & 49.7 \\
    \textbf{Ensemble (remove softmax)} & 86.4 & 48.7 \\
    \bottomrule
    \end{tabular}
\end{table}

\subsection{Ensemble Attacks}

In the experiments reported in the main body of the paper, we treated the entire ensemble as a single differentiable model for performing adversarial evaluations. Here, we also consider a number of alternate attack strategies, which do not require differentiating through a softmax layer. First, we considered a number of ``transfer'' attacks, where we treat all $4$ models as separate models, and optimize with respect to either a randomly selected model at each iteration \cite{athalye2018synthesizing}, or the mean or maximum adversarial loss across the $4$ networks. We found that using the mean adversarial loss was the most effective (slightly better than a randomly selected model, for sufficiently high numbers of iterations, and significantly better than max), so we report numbers for this attack. We also experimented with clipping the loss associated with each individual network, but found this weakened the attacks.

\begin{figure}[h]
\begin{subfigure}[b]{\linewidth}
    \centering
    \includegraphics[width=0.9\linewidth,trim={.5cm 0cm 0cm .5cm},clip]{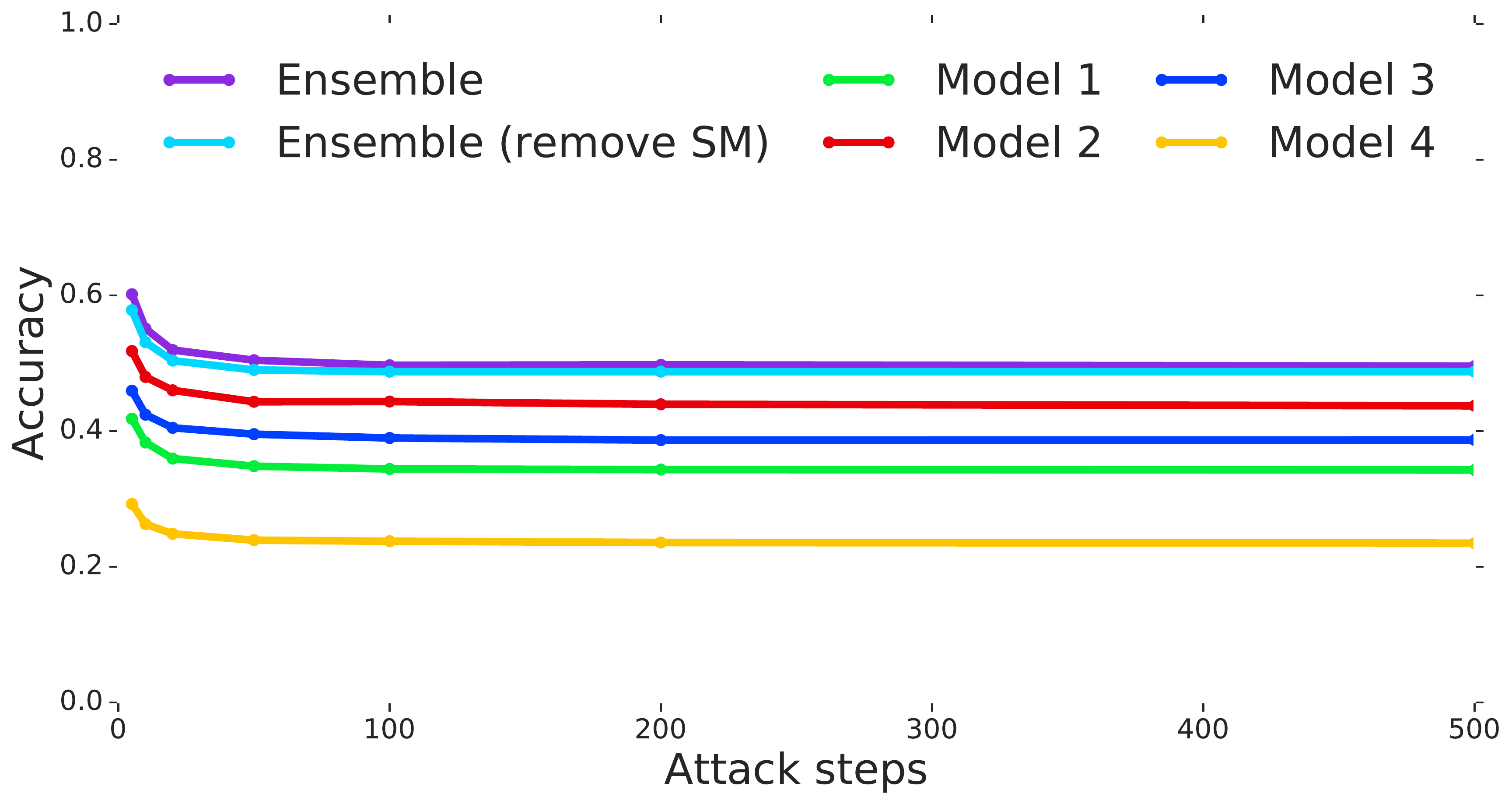}
    \caption{Accuracy under PGD attack as a function of the number of attack steps.}
\end{subfigure}
\caption{Additional Evaluation Curves}
\label{fig:appendix_plots}
\end{figure}

We show the strength of this attack for varying numbers of iterations in Figure \ref{fig:appendix_plots}, which indicates attack strength plateaus after $100$ iterations. Table \ref{tab:eval_tables_extra} reports accuracy against strongest attack.

Finally, we consider an alternative ensembling model, where we use the same parameters for \textbf{Ensemble4Adv}, but average in logit space rather than probability space. This is referred to as ``Ensemble (remove softmax)'' in Figure \ref{fig:appendix_plots} and Table \ref{tab:eval_tables_extra}. This evaluation provides the greatest confidence the model is not obscured, as the model does not contain a softmax, and the attacked model matches the inference procedure exactly. However, as the inference procedure does not match the procedure used for training, we expect the accuracy for this model will be somewhat worse than the \textbf{Ensemble4Adv} model. If the PGD attack against the no-softmax ensemble model is near-optimal, the true adversarial accuracy of \textbf{Ensemble4Adv} is unlikely to be much less than $48.7\%$.

\end{document}